\documentclass[preprint,12pt]{elsarticle}

\usepackage{amssymb}
\usepackage[T1]{fontenc}
\usepackage[utf8]{inputenc}
\usepackage{amssymb}
\usepackage{wasysym}
\usepackage{tikz}
\usepackage{comment}
\usepackage{color}
\usepackage{booktabs}
\usepackage[normalem]{ulem}
\usepackage{listings}     
\usepackage{lstautogobble}  
\usepackage{color}          
\usepackage{zi4}            
\usepackage{subcaption}
\usepackage{url}
\usepackage{listings}
\definecolor{bluekeywords}{rgb}{0.13, 0.13, 1}
\definecolor{greencomments}{rgb}{0, 0.5, 0}
\definecolor{redstrings}{rgb}{0.9, 0, 0}
\definecolor{graynumbers}{rgb}{0.5, 0.5, 0.5}
\usepackage{mdframed} 
\usepackage{multirow}

\usepackage{tabularx}
\newcolumntype{C}{>{\centering\arraybackslash}X}
\newcolumntype{R}{>{\raggedleft\arraybackslash}X}
\newcolumntype{L}{>{\raggedright\arraybackslash}X}

\definecolor{red}{HTML}{FD0100}
\definecolor{orange}{HTML}{F76915}
\definecolor{yellow}{HTML}{EEDE04}
\definecolor{green}{HTML}{A0D636}
\definecolor{teal}{HTML}{2FA236}
\definecolor{blue}{HTML}{333ED4}

\usepackage{makecell}

\journal{Neurocomputing}

\begin{document}

\begin{frontmatter}

\title{\emph{problexity} --- an open-source \emph{Python} library for binary classification problem complexity assessment}

\author{Joanna Komorniczak and Pawel Ksieniewicz}

\address{Department of Systems and Computer Networks,\\ Faculty of Information and Communication Technology,\\Wrocław University of Science and Technology,\\Wybrzeże Wyspiańskiego 27, 50-370 Wrocław, Poland}

\begin{abstract}
    The classification problem's complexity assessment is an essential element of many topics in the supervised learning domain. It plays a significant role in meta-learning -- becoming the basis for determining meta-attributes or multi-criteria optimization -- allowing the evaluation of the training set resampling without needing to rebuild the recognition model. The tools currently available for the academic community, which would enable the calculation of problem complexity measures, are available only as libraries of the \emph{C++} and \emph{R} languages. This paper describes the software module that allows for the estimation of 22 complexity measures for the Python language -- compatible with the \emph{scikit-learn} programming interface -- allowing for the implementation of research using them in the most popular programming environment of the \emph{machine learning} community.
\end{abstract}

\begin{keyword}
    Problem complexity \sep Classification \sep Python
\end{keyword}

\end{frontmatter}

\section{Motivation and significance}

\noindent
Proper evaluation of the algorithms and methods proposed for solving the pattern classification task, in accordance with good practices adopted in the machine learning research environment~\cite{stapor2021design}, requires extensive experiments performed on a large pool of appropriately diverse data sets~\cite{hoffmann2019benchmarking}. In a typical approach to designing an experimental evaluation procedure, the usefulness of a selected group of problems is most often assessed by basic reporting of the sets dimensionality, the number of examples describing them, the number of problem classes~\cite{sotoca2005review} and their prior distribution -- in the case of non-uniformity of their representation~\cite{fernandez2018learning}. However, it is necessary to remember that these are only simple measures, briefly describing the problem difficulty without giving the researcher sufficient insight into the actual complexity of the task. A true complexity of a problem is contained not only in the basic characteristics of problem space or class imbalance but also in problem-specific distribution, understood as its linearity, neighborhood characteristic, geometrical and topological complexity, and feature dependency~\cite{ho2002complexity}.

The classification task is the main issue of supervised machine learning, finding its applications in almost every branch of life and science, starting from economics, through epidemiology and medicine, to machine vision and predictive maintenance systems~\cite{soofi2017classification}. Such a variety of undertaken problems leads to a plethora of various difficulties, expressed in a multiplicity of class-building clusters, a significant imbalance ratio, or significant class overlap, to enumerate a few. The considered complexity metrics demonstrate the potential for assessing the diversity of collected data sets which were precisely described and spanned over the taxonomy by Lorena et al.~\cite{lorena2019complex}.

The computations of the problem complexity do not only find their application in the proper selection of benchmark datasets for the experimental evaluation. Their application in \emph{meta-learning} solutions has been particularly popular in recent years~\cite{vanschoren2018meta}, being one of the primary sources of meta-features for problem identification. Such an approach allows for the induction of task representations~\cite{pmlr-v140-meskhi21a} and the automation of the deep neural networks structure configuration~\cite{Konuk_2019_ICCV}. The measures of problem complexity are also used in geospatial data -- in filtering the predictor noise~\cite{guillon2020machine} -- or in the difficulty metrics dedicated to spectral data~\cite{Branchaud-Charron_2019_CVPR}. Preliminary works assessing the usefulness of such measures in data stream processing are also present in the literature~\cite{ellis2021characterisation}.

All the above-mentioned examples concern research from the narrow span of the last few years, in which researchers carried out elements of problem complexity analysis with the use of two problem-complexity libraries currently available to the scientific community.

The first library -- \textit{DCoL} -- was developed in 2010 by the team of \emph{Universitat Ramon Llull} and \emph{Bell Laboratories}~\cite{orriols2010documentation} and is an implementation of 14 basic complexity metrics for the \emph{C++} language. In 2017 its source code was made available on the GitHub\footnote{\url{https://github.com/nmacia/dcol}}.

A second library, \textit{ECoL}, published as supporting software for Lorena et al. ~\cite{lorena2019complex} publication, has been developed since September 2016 and is an implementation of 22 metrics in \emph{R} language, currently being the most comprehensive solution for this type available to the scientific community. It reached its current version (\verb|0.3.0|) in December 2020. Its entire version history is available in public GitHub repository\footnote{\url{https://github.com/lpfgarcia/ECoL/}}.

It is important to emphasize that in recent years, the \emph{Python} programming language has started playing a much more significant role in the development of machine learning methods than any other experimental environment~\cite{nguyen2019machine}.  

This publication presents a \emph{problexity} library, containing the implementation of 22 problem complexity measures divided to six categories: (\emph{i}) feature-based, (\emph{ii}) linearity, (\emph{iii}) neighborhood, (\emph{iv}) network, (\emph{v}) dimensionality, and (\emph{vi}) class imbalance as well as a \emph{ComplexityCalculator} class, introducing additional utilities facilitating research and enabling simple expansion of the module with additional metrics. By proposing this library, new methods considering the classification complexity can be developed, which will further impact the evolution of machine learning algorithms both in the meta-learning field and in other research applications from recent years.

The measures of the problem complexity can be used as a substitute criterion in optimization tasks. We can expect that the quality of the classification will depend on the problem's difficulty expressed in measures of complexity. Testing the classifier's ability to recognize objects, often using cross-validation and induction algorithms with computational overhead significantly larger than measure computation, will be time consuming. We can potentially speed up the optimization process by problem complexity assessment as an alternative criterion.

\section{Software description}

\noindent
This chapter contains the software description of the library. It will present the package structure, a minimal processing example, and an exemplary analysis of the results. The chapter will illustrate the implemented measures of problem complexity and the \emph{ComplecityCalculator} model.

\subsection{Software Architecture}

\noindent
The library consists of two main elements -- the \emph{measures} submodule and the \emph{ComplexityCalculator} class. Within the measures, six categories are distinguished:

\begin{itemize}
    \item \emph{Feature-based} containing $F1$, $F1v$, $F2$, $F3$, and $F4$ measures,
    \item \emph{Linearity} containing  $L1$, $L2$, and $L3$ measures,
    \item \emph{Neighborhood} with $N1$, $N2$, $N3$, $N4$, $T1$, and $LSC$ measures,
    \item \emph{Network} containing $density$, $ClsCoef$, and $Hubs$ measures,
    \item \emph{Dimensionality} containing $T2$, $T3$, and $T4$ measures,
    \item \emph{Class imbalance} containing $C1$ and $C2$ measures.
\end{itemize}

The \emph{ComplexityCalculator} module enables the computation and analysis of data sets in the context of the difficulty of the classification task. It offers a set of methods that allows calculating the measure values and presenting the result as a single score, report, or illustrative graph.

\begin{figure}[!ht]
    \centering
    \resizebox{13cm}{!}{    \usetikzlibrary{arrows}
    \begin{tikzpicture}[
        mainblock/.style={draw, rounded corners=.5cm, text width=12cm, minimum height=3cm, minimum width=18.5cm, align = center, thick},
        mainblock2/.style={draw, rounded corners=.5cm, text width=12cm, minimum height=3cm, minimum width=14.5cm, align = center, thick},
        mainblock3/.style={draw, rounded corners=.5cm, text width=12cm, minimum height=3cm, minimum width=13.5cm, align = center, thick},
        midblock/.style={draw, text width=6.25cm, minimum height=.75cm, minimum width=7cm, align = left, thick},
        header/.style={align = center, fill=white},
        header2/.style={align = left, fill=white, anchor=west},
        q/.style={white!25!black, thick},
    ]
        \draw[step=1, white, thin, dotted] (0.0,0.0) grid (20,15);
        
        \node[mainblock, minimum height=6cm] at (10,14) () {};
        \node[header] at(10, 17) {\bfseries ComplexityCalculator};
        
        \node[midblock, minimum height=5.5cm] at (6,14) () {\verb|fit(X, y)|\\{\footnotesize Calculates metrics for given dataset.} };
        
        \node[midblock, minimum height=1.5cm] at (14,16) () {\verb|report()|\\ {\footnotesize Returns report of problem complexity.}};
        \node[midblock, minimum height=1.5cm] at (14,14) () {\verb|score()|\\ {\footnotesize Returns integrated score of problem complexity.}};
        \node[midblock, minimum height=1.5cm] at (14,12) () {\verb|plot()|\\ {\footnotesize Draws \verb|matplotlib| figure illustrating problem complexity.}};

        \draw[->, q, dashed] (9.5, 16) -- (10.4, 16);
        \draw[->, q, dashed] (9.5, 14) -- (10.4, 14);
        \draw[->, q, dashed] (9.5, 12) -- (10.4, 12);
        
        \node[mainblock2, minimum height=9.5cm] at (12,5.25) () {};
        \node[header] at(12, 10) {\bfseries Measures};
        
        \node[mainblock3, minimum height=1cm] at (12,9) () {\emph{f1(), f1v(), f2(), f3(), f4()}};
        \node[header2] at(5.5, 9.5)  {\textbf{\emph{feature based}}};
        
        \node[mainblock3, minimum height=1cm] at (12,7.5) () {\emph{l1(), l2(), l3()}};
        \node[header2] at(5.5, 8)  {\textbf{\emph{linearity}}};
        
        \node[mainblock3, minimum height=1cm] at (12,6) () {\emph{n1(), n2(), n3(), n4(), t1(), lsc()}};
        \node[header2] at(5.5, 6.5)  {\textbf{\emph{neighborhood}}};
        
        \node[mainblock3, minimum height=1cm] at (12,4.5) () {\emph{density(), clsCoef(), hubs()}};
        \node[header2] at(5.5, 5)  {\textbf{\emph{network}}};
        
        \node[mainblock3, minimum height=1cm] at (12,3) () {\emph{t2(), t3(), t4()}};
        \node[header2] at(5.5, 3.5)  {\textbf{\emph{dimensionality}}};
        
        \node[mainblock3, minimum height=1cm] at (12,1.5) () {\emph{c1(), c2()}};
        \node[header2] at(5.5, 2)  {\textbf{\emph{class imbalance}}};
        
        \draw[->, q] (4.75, 5.25) -- (3, 5.25) -- (3,10.9);

    \end{tikzpicture}}
    \caption{Overall schema of the software architecture}
    \label{fig:architecture}
\end{figure}
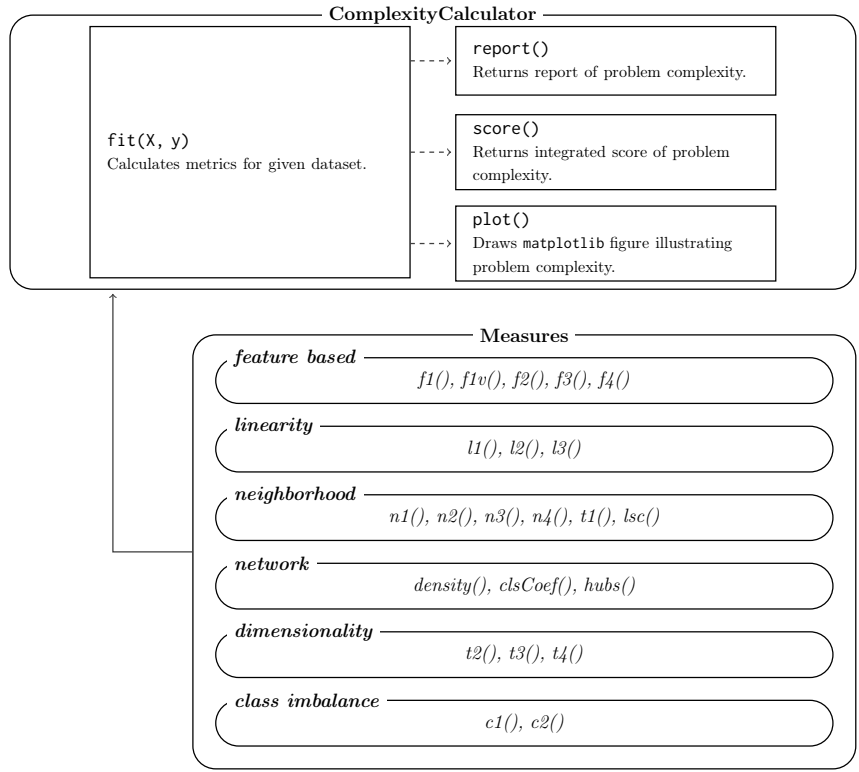

\subsection{Measures}

\noindent
The package divides measures into six categories, introduced in the publication by Lorena et al. \cite{lorena2019complex}. The measures return $0$ or a value close to $0$ for simple problems and values close to $1$ for complex problems. The only measures not limited to 1 are $T2$ and $T3$ from the \emph{dimensionality} category. 
Four implemented measures ($L1$, $L2$, $L3$, $N4$) are non-deterministic; therefore, subsequent calculations on the same data can yield varying results. In the case of measures from the Linear category, this behavior results from Linear SVM classifier optimization, whose weights are initialized randomly. In the case of L3 and N4 measures, the generation of synthetic instances in a randomized manner makes the calculations non-deterministic.

\subsubsection{Feature-based measures} 
\noindent
The measures describe the ability of features to separate classes in the classification problem. They analyze features separately or evaluate how attributes work together.

\begin{itemize}
    \item[$F1$] \textbf{Maximum Fisher's discriminant ratio}.

    The measure describes the overlap of feature values in each class. The inverse of the original formulation, taking into account the most significant discriminant ratio, is taken into account, the same as in \emph{ECoL} package.
    
    \item[$F1v$] \textbf{Directional vector maximum Fisher's discriminant ratio}.

    The measure computes projection that maximizes class separation by directional Fisher's criterion.

    \item[$F2$] \textbf{Volume of overlapping region}

    The measure describes the overlap of the feature values within the classes. It is determined by the minimum and maximum values of features in each class. The overlap is then calculated and normalized by the range of values in each class.

    \item[$F3$] \textbf{Maximum individual feature efficiency}

    The measure describes the efficiency of each feature in the separation of classes. It considers the maximum value among all features. The equation proposed by Lorena et al. \cite{lorena2019complex} has been slightly modified to obtain a maximum complexity value of 1 in case all instances of separate classes overlap.

    \item[$F4$] \textbf{Collective feature efficiency}

    The measure describes the features synergy. The instances separated by the most discriminant attribute that was not used already are excluded from further analysis. The process continues until all instances are classified, or all features are used. The measure is calculated according to the number of instances in the overlapping region and the total number of samples.

\end{itemize}

\subsubsection{Linearity measures} 
\noindent
The measures evaluate the level of problem class linear separation. The measures use Linear Support Vector Machines (\textsc{svm}) classifier.

\begin{itemize}
    \item[$L1$] \textbf{Sum of the error distance by linear programming}

    The measure calculates the distance of incorrectly classified samples from the \textsc{svm} hyperplane.

    \item[$L2$] \textbf{Error rate of linear classifier}

    The measure is described by the error rate of the Linear \textsc{svm} classifier within the dataset.

    \item[$L3$] \textbf{Non-linearity of linear classifier}
    The measure is described by the classifier's error rate on synthesized points of the dataset. The synthetic points are obtained by linearly interpolating instances of each class. The class of original examples determines the label of an augmented point, and the number of artificial points is equal to the original dataset size.

\end{itemize}

\subsubsection{Neighborhood measures} 
\noindent
The measures analyze the neighborhood of instances in a feature space. Neighbors of each sample are established based on the distance between problem instances.

\begin{itemize}
    \item[$N1$] \textbf{Fraction of borderline points}

    The \emph{Minimum Spanning Three} is generated over input instances in order to obtain this measure. The value is computed by calculating the number of edges in the \textsc{mst} between examples of different classes over a total number of samples. 
    
    \item[$N2$] \textbf{Ratio of intra/extra class NN distance}

    The measure depends on the distances of each problem instance to its nearest neighbor of the same class and the distance to the nearest neighbor of a different class. According to the proportions of those values, the final value is calculated.

    \item[$N3$] \textbf{Error rate of NN classifier}

    The measure is determined by the error rate of the One Nearest Neighbor Classifier in the Leave One Out evaluation protocol.

    \item[$N4$] \textbf{Calculates the Non-linearity of NN classifier (N4) metric}

    The measure is determined by the error rate of the k-Nearest Neighbor Classifier on synthetic points, generated by linearly interpolating original instances. The classifier is fitted on original points and evaluated on synthetic instances.

    \item[$T1$] \textbf{Fraction of hyperspheres covering data}

    The measure is defined by the number of hyperspheres needed to cover the data divided by a number of instances. First, a hypersphere is generated for each problem sample. A sample lies in the center of the hypersphere. Its radius is dependent on the distance to the instance of another class. The hyperspheres are eliminated if a different one already covers the center instance. The elimination starts from the hyperspheres with the largest radius and continues to the ones with a smaller radius. The hyperspheres that were not eliminated are taken into account during the calculation of complexity.

    \item[$LSC$] \textbf{Local set average cardinality (LSC)}

    The measure is dependent on the distances between instances and the distances to the instances' nearest enemies -- the nearest sample of the opposite class. The number of cases that lie closer to the sample than its closest enemy is considered during the calculation. 

\end{itemize}

\subsubsection{Network measures} 
\noindent
The measures consider the instances as the vertices of the graph. All measures of this category generate an epsilon-Nearest Neighbours graph. The epsilon value is set to 0.15, same as in the \emph{ECoL} package. The edges are selected based on the Gower distance between samples, normalized to the range between 0 and 1. The edge is placed between the points if a normalized Gower distance is smaller than 0.15. Edges between instances of distinct classes are removed.

\begin{itemize}
    \item[$density$] \textbf{Density metric}

    The measure calculates the number of edges in the final graph divided by the total possible number of edges.

    \item[$clsCoef$] \textbf{Clustering Coefficient metric}

    For the purpose of obtaining this measure, the neighborhood of each vertex is calculated, i.e., the instances directly connected to it. Then, the number of edges between the sample's neighbors is calculated and divided by the maximum possible number of edges between them. The final measure is calculated based on the neighborhood of each point in the dataset.

    \item[$hubs$] \textbf{Hubs metric}

    For the purpose of obtaining this measure, the neighborhood of each vertex is obtained. The measure scores each sample by the number of connections to neighbors, weighted by the number of connections the neighbors have.

\end{itemize}

\subsubsection{Dimensionality measures} 
\noindent
The measures analyze the relation between the number of features and the number of instances in the dataset.

\begin{itemize}
    \item[$T2$] \textbf{Average number of features per dimension} 

    For the purpose of obtaining this measure, the number of dimensions describing the dataset is divided by the number of instances.
    
    \item[$T3$] \textbf{Average number of PCA dimensions per points}

    To obtain this measure, first, the number of PCA components needed to represent 95\% of data variability is calculated. Then, the value is divided by the instance number in the dataset.

    \item[$T4$] \textbf{Ration of the PCA dimension to the original dimension}

    To obtain this measure, the number of PCA components needed to represent 95\% of data variability is divided by the original number of dimensions. This measure describes the proportion of relevant dimensions in the dataset.
    
\end{itemize}

\subsubsection{Class imbalance measures} 
\noindent
The Class Imbalance measures evaluate the dataset based on the degree of data imbalance.

\begin{itemize}
    \item[$C1$] \textbf{Entropy of Class Proportions}
    The measure is obtained based on the proportion of each class's samples divided by the total number of samples. 

    \item[$C2$] \textbf{Imbalance Ratio}

    The measure is obtained based on the proportion of each class's samples divided by a number of opposite class samples. 
    
\end{itemize}

\subsection{ComplexityCalculator}
\noindent
The library introduces the \emph{ComplexityCalculator} class to facilitate the use of measure implementation. Its objects are initialized with a list of measures and an optional list of (\emph{a}) category colors for visualization purposes, and (\emph{b}) a dictionary indicating the number of measures in a given category, which are necessary only in case of non-default collection of measures.

By default, the module will analyze all 22 metrics of the \emph{measures} module. Executing the \verb|fit()| method, which takes a set of features $X$ and a set of labels $y$ as an argument, will calculate the values of the metrics. 

The obtained measures' values can be accessed as a single value using the \verb|score()| method, optionally taking a vector of weights as a parameter. The vector length has to correspond to the number of analyzed measures. By default, each measure has an equal weight, which means that executing the \verb|score()| method will return the arithmetic mean of measured complexities.

The \verb|report()| method provides a more detailed description of the classification problem. The method returns a dictionary containing a summary of each metric value, their arithmetic mean, and other data set characteristics, such as the number of samples, dimensionality, labels, and the number of classes with a prior probability.

The values can also be presented in the form of a graph. Executing the plot method returns a chart that illustrates the value of each measure from respective categories as well as the default score of the problem.

\subsection{Minimal processing example}
\noindent
The \emph{problexity} module is open Python software released under the \emph{GPL-3.0} license and versioned in the public \emph{Python Package Index} (PyPI) repository. Therefore, it can be easily obtained with the \emph{pip} package installer with the command:

\begin{lstlisting}[language=bash,numbers=none]
> pip install problexity
\end{lstlisting}

To enable the possibility to modify the measures provided by \verb|problexity| or in case of necessity to expand it with functions that it does not yet include, it is also possible to install the module directly from the source code. If any modifications are introduced, they propagate to the module currently available to the environment.

\begin{lstlisting}[language=bash,numbers=none]
> git clone https://github.com/w4k2/problexity.git
> cd problexity
> make install
\end{lstlisting}

The \verb|problexity| module is imported in the standard Python fashion. At~the same time, for the convenience of implementation, the authors recommend importing it under the \verb|px| alias:

\begin{lstlisting}[language=Python]
# Importing problexity
import problexity as px
\end{lstlisting}

The library is equipped with the \emph{ComplexityCalculator} calculator, which serves as the basic tool for establishing metrics. The following code presents an example of the generation of a synthetic data set -- typical for the \emph{scikit-learn} module -- and the determination of the value of measures by fitting the complexity model in accordance with the standard API adopted for \emph{scikit-learn} estimators:

\begin{lstlisting}[language=Python]
# Loading benchmark dataset from scikit-learn
from sklearn.datasets import load_breast_cancer
X, y = load_breast_cancer(return_X_y=True)

# Initialize CoplexityCalculator with default parametrization
cc = px.ComplexityCalculator()

# Fit model with data
cc.fit(X,y)
\end{lstlisting}

As the $L1$, $L2$ and $L3$ measures use the recommended \emph{LinearSVC} implementation from the \verb|svm| module of the \emph{scikit-learn} package in their calculations, the warning "\emph{ConvergenceWarning: Liblinear failed to converge, increase the number of iterations.}" might occur. It is not a problem for the metric calculation -- only indicating the lack of linear problem separability.

The complexity calculator object stores a list of all estimated measures that can be read by the model's \verb|complexity| attribute:

\begin{lstlisting}[language=Python]
cc.complexity
\end{lstlisting}
\vspace{-.5em}\begin{mdframed}[backgroundcolor=black!5, leftmargin=1, rightmargin=1, innerleftmargin=5, innertopmargin=5,innerbottommargin=5, outerlinewidth=.5, linecolor=white]
\begin{lstlisting}[numbers=none,basicstyle=\color{black!75}\ttfamily\footnotesize,
framexleftmargin = 1.5em]
[0.227 0.064 0.000 0.478 0.012 0.225 0.070 0.042 0.043 0.296 0.084
 0.025 0.178 0.912 0.741 0.268 0.569 0.053 0.002 0.033 0.047 0.122]
\end{lstlisting}
\end{mdframed}

They appear in the list in the same order as the declarations of the used metrics, which can also be obtained from the hidden method \verb|_metrics()|:

\begin{lstlisting}[language=Python]
cc._metrics()
\end{lstlisting}
\vspace{-.5em}\begin{mdframed}[backgroundcolor=black!5, leftmargin=1, rightmargin=1, innerleftmargin=5, innertopmargin=5,innerbottommargin=5, outerlinewidth=.5, linecolor=white]
\begin{lstlisting}[numbers=none,basicstyle=\color{black!75}\ttfamily\footnotesize,
framexleftmargin = 1.5em]
['f1', 'f1v', 'f2', 'f3', 'f4', 'l1', 'l2', 'l3', 'n1', 'n2', 'n3', 
 'n4', 't1', 'lsc', 'density', 'clsCoef', 'hubs', 't2', 't3', 't4', 
 'c1', 'c2']
\end{lstlisting}
\end{mdframed}

The problem difficulty score can also be obtained as a single scalar measure, which is the arithmetic mean of all measures used in the calculation:

\begin{lstlisting}[language=Python]
cc.score()
\end{lstlisting}
\vspace{-.5em}\begin{mdframed}[backgroundcolor=black!5, leftmargin=1, rightmargin=1, innerleftmargin=5, innertopmargin=5,innerbottommargin=5, outerlinewidth=.5, linecolor=white]
\begin{lstlisting}[numbers=none,basicstyle=\color{black!75}\ttfamily\footnotesize,
framexleftmargin = 1.5em]
0.203
\end{lstlisting}
\end{mdframed}

The \emph{problexity} module, in addition to raw data output, also provides two standard representations of problem analysis. The first is a report in the form of a dictionary presenting the number of patterns (\verb|n_samples|), attributes (\verb|n_features|), classes (\verb|classes|), their prior distribution (\verb|prior_probability|), average metric (\verb|score|) and all member metrics (\verb|complexities|), which can be obtained using the model's \verb|report()| method:

\begin{lstlisting}[language=Python]
cc.report()
\end{lstlisting}
\vspace{-.5em}\begin{mdframed}[backgroundcolor=black!5, leftmargin=1, rightmargin=1, innerleftmargin=5, innertopmargin=5,innerbottommargin=5, outerlinewidth=.5, linecolor=white]
\begin{lstlisting}[language=Python, numbers=none,basicstyle=\color{black!75}\ttfamily\footnotesize,
framexleftmargin = 1.5em]
{
    'n_samples': 569, 
    'n_features': 30, 
    'n_classes': 2, 
    'classes': array([0, 1]), 
    'prior_probability': array([0.373, 0.627]), 
    'score': 0.214, 
    'complexities': 
    {
        'f1': 0.227, 'f1v': 0.064, 'f2': 0.001, 'f3': 0.478, 'f4': 0.012, 
        'l1': 0.433, 'l2' : 0.069, 'l3': 0.049, 'n1': 0.043, 'n2': 0.296, 
        'n3': 0.084, 'n4' : 0.039, 't1': 0.178, 't2': 0.053, 't3': 0.002, 
        't4': 0.033, 'c1' : 0.047, 'c2': 0.122,
        'lsc': 0.912, 'density': 0.741, 'clsCoef': 0.268, 'hubs': 0.569
    }
}
\end{lstlisting}
\end{mdframed}

The second form of reporting is a graph which, in the polar projection, collates all metrics, grouped into categories using color codes:

\begin{itemize}
    \item[\bfseries\color{red}\textsc{red}] -- feature based measures,
    \item[\bfseries\color{orange}\textsc{orange}] -- linearity measures,
    \item[\bfseries\color{yellow}\textsc{yellow}] -- neighborhood measures,
    \item[\bfseries\color{green}\textsc{green}] -- network measures,
    \item[\bfseries\color{teal}\textsc{teal}] -- dimensionality measures,
    \item[\bfseries\color{blue}\textsc{blue}] -- class imbalance measures.
\end{itemize}

\begin{figure}[!ht]
    \centering
    \includegraphics[width=.75\textwidth]{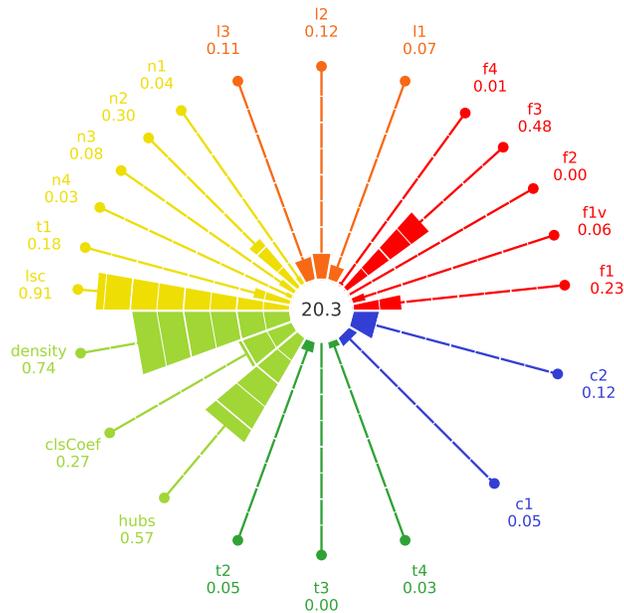}
    \caption{Exemplary complexity graph generated by \emph{problexity} module}
    \label{fig:example_radar}
\end{figure}

Each problem difficulty category occupies the same graph area, meaning that contexts that are less numerous in metrics (class imbalance) are not dominated in this presentation by categories described by many metrics (neighborhood). The illustration is built with the standard tools of the \verb|matplotlib| module as a subplot of a figure and can be generated with the following source code:

\begin{lstlisting}[language=Python]
# Import matplotlib
import matplotlib.pyplot as plt

# Prepare figure
fig = plt.figure(figsize=(7,7))

# Generate plot describing the dataset
cc.plot(fig, (1,1,1))
\end{lstlisting}

An example of a complexity graph is shown in the Figure~\ref{fig:example_radar}.

\section{Comparison with available modules}
\noindent
The juxtaposition in Table \ref{tab:comparisonone} presents a comparison of the available libraries analyzing the difficulty of classification problems. The columns successively present the \emph{ECoL}, \emph{DCoL}, and the \emph{problexity} libraries. The rows show the categories of the complexity measures. The values in the cells indicate the number of available metrics compared to the number described in a publication by Lorena et al. \cite{lorena2019complex}. 

Since \emph{DCoL} was the earliest implemented library of the analyzed ones, it contains the fewest measures. The \emph{ECoL} and \emph{problexity} libraries are based on the same publication, so the number of measures in each category matches. The Table also shows the availability of utilities offered by the libraries and basic information describing them. All of the compared packages offer the possibility of generating a report containing a summary of the values of selected measures. In addition, the \emph{problexity} package includes a method allowing to represent selected measures as a single value and as well contains a tool for graphically presenting the measures.

\begin{table}[!htb]
\footnotesize
    \centering
    \caption{Comparison of measures and utilities available in \emph{ECoL}, \emph{DCoL} and \emph{problexity}}
    
\begin{tabularx}{\textwidth}{@{}lLccc@{}}
\toprule
\textsc{area} & \textsc{functionality} &   \emph{ECoL}           &   \emph{DCoL}           &   \emph{problexity}            \\ \midrule
\emph{Measures}       & Feature-based                           & 5/5          & 5/5          & 5/5                 \\
& Linearity   & 3/3          & 3/3          & 3/3                 \\
& Neighborhood & 5/5          & 5/5          & 5/5                 \\
& Network     & 3/3          & 0/3          & 3/3                 \\
& Dimensionality  & 3/3          & 1/3          & 3/3                 \\
& Class imbalance  & 2/2          & 0/2          & 2/2                 \\ \midrule
\emph{Utility Module} & Score       &              &              &  {\small\checkmark} \\
    & Report      & {\small\checkmark} & {\small\checkmark} &   {\small\checkmark} \\
    & Plot        & &  & {\small\checkmark} \\\midrule

\emph{Basic information} 
& Language & R & C++ & Python\\
& Current version & 0.3.0 & --- & 0.3.2\\
\bottomrule
\end{tabularx}

    \label{tab:comparisonone}
\end{table}

\section{Impact}
\noindent
In recent years, the complexity measures for classification problems have gained particular interest in the scientific community. Their most common use is the construction of meta-attributes (features describing sets~\cite{rivolli2018characterizing}) to automate the selection of processing flows typical of the meta-learning topic~\cite{vanschoren2018meta}. An interesting trend here is, in particular, the construction of abstract representations of recognition tasks~\cite{pmlr-v140-meskhi21a}, which allow for an initial generalization of the problem under consideration~\cite{rivolli2022meta}, allowing for a significant reduction of the time necessary to select the optimal classification model for a given task~\cite{Konuk_2019_ICCV}. 

An alternative field of use is the classification of difficult data, with particular emphasis on multidimensional discrete signals -- for example -- in the form of multispectral and geospatial problems~\cite{Branchaud-Charron_2019_CVPR}. On the one hand, the measures of complexity allow for agnostic estimation of the correct structure of the objects in the predictor's action space~\cite{garcia2015effect}. On the other hand, they are tools useful in filtering its noise~\cite{guillon2020machine}. The same agnostic characteristics show a particular potential in the processing of imbalanced data~\cite{lee2021efficient}, allowing the quality of the proposed resampling to be assessed without having to rebuild the recognition model~\cite{barella2018data}. 

As with imbalanced data, complexity measures also find their application in processing data streams~\cite{ellis2021characterisation}. By demonstrating the relationship between the quality of the recognition model and some currently available measures, it is possible to use them as a proxy-classifier, which allows for a significant reduction in the time of reviewing available solutions when using any optimization methods~\cite{cai2019classification}.

The field of classification problem complexity is still very active, not only in applications described above but also in the proposals of new measures that appear each year. The newly proposed measures allow the assessment of other processing contexts such as category learning~\cite{rosedahl2019difficulty}, rule-based dissociations~\cite{ashby2020dissociations}, or lost points identification~\cite{lancho2021complexity}, to enumerate a few.

\section{Conclusions}
\noindent
This paper presents the \emph{problexity} library for \emph{Python} programming language. The library contains measures for assessing the complexity of binary classification problems. Twenty-two evaluation measures of the problem complexity have been implemented in the following categories: feature-based, linearity, neighborhood, network, dimensionality, and class imbalance. Additionally, the library incorporates the \emph{ComplexityCalculator} module, which provides additional tools for analyzing classification data sets.

The library was created to fill the gap related to the lack of accessible measures for assessing the complexity of the classification problem in Python. Increasing the availability of complexity assessment methods and creating a tool for exploring them will allow for a more detailed analysis of data sets' characteristics, potentially impacting the development of new machine learning methods.

The current version of the package offers a set of measures adapted to binary classification datasets, which are the most frequent objective of machine learning applications. The intent of future versions is for the library to include measures adapted to multiclass problems. 
As we pointed out in the impact section, in the field of data complexity evaluation, new measures are being proposed. The package will be maintained to contain other measures of classification complexity and possibly extended with regression complexity evaluation measures. Further works will focus on adapting the library to analyze data streams in the context of data difficulties. Finally, the \emph{problexity} library will continue to be used in research studies of the classification problem complexity employment.

\section*{Acknowledgements}
\noindent
This work was supported by the Polish National Science Centre under the grant No. 2019/35/B/ST6/0442 as well as by the statutory funds of the Department of Systems and Computer Networks, Faculty of Information and Communication Technology, Wroclaw University of Science and Technology.

\bibliographystyle{elsarticle-num} 
\bibliography{bibliography}

\clearpage
\section*{Required Metadata}
\label{}

\section*{Current executable software version}
\label{}

\begin{table}[!ht]
\caption{Code metadata (mandatory)}
\begin{tabular}{|l|p{6.5cm}|p{6.5cm}|}
\hline
\textbf{Nr.} & \textbf{Code metadata description} & \textbf{Please fill in this column} \\
\hline
C1 & Current code version & 0.3.2 \\
\hline
C2 & Permanent link to code/repository used for this code version & $https://github.com/w4k2/problexity$ \\
\hline
C3 & Legal Code License & GPL-3.0 \\
\hline
C4 & Code versioning system used & git \\
\hline
C5 & Software code languages, tools, and services used & python \\
\hline
C6 & Compilation requirements, operating environments \& dependencies & \\
\hline
C7 & If available Link to developer documentation/manual &  $https://problexity.readthedocs.io$ \\
\hline
C8 & Support email for questions & $joanna.komorniczak@pwr.edu.pl$\\
\hline
\end{tabular}
\end{table}

\end{document}